\documentclass[letterpaper,english]{article}
\usepackage[T1]{fontenc}
\usepackage[latin9]{inputenc}
\usepackage{color}
\usepackage{babel}
\usepackage{array}
\usepackage{calc}
\usepackage{multirow}
\usepackage{amsmath}
\usepackage{graphicx}
\usepackage[unicode=true,pdfusetitle,
 bookmarks=true,bookmarksnumbered=false,bookmarksopen=false,
 breaklinks=true,pdfborder={0 0 1},backref=false,colorlinks=true]
 {hyperref}

\makeatletter

\pdfpageheight\paperheight
\pdfpagewidth\paperwidth

\providecommand{\tabularnewline}{\\}

\usepackage{aaai19}

\usepackage{times}
\usepackage{helvet}
\usepackage{courier}
\usepackage{url}
\usepackage{graphicx}
\usepackage{xspace}
\DeclareRobustCommand\onedot{\futurelet\@let@token\@onedot}
\def\@onedot{\ifx\@let@token.\else.\null\fi\xspace}

\def\eg{\emph{e.g}\onedot} 
\def\ie{\emph{i.e}\onedot}

\def\etal{\emph{et al}\onedot}

\@ifundefined{showcaptionsetup}{}{%
 \PassOptionsToPackage{caption=false}{subfig}}
\usepackage{subfig}
\makeatother

\begin{document}

\global\long\def\product{\cdot}


\global\long\def\centerTarget{C}

\global\long\def\ourModel{\text{HitNet}}


\global\long\def\cardinalityOfClasses{K}

\global\long\def\classIndex{k}


\global\long\def\loss{L}

\global\long\def\output{y}

\global\long\def\true#1{#1_{\text{true}}}

\global\long\def\indexedTrue#1#2{#1_{\text{true},\,#2}}

\global\long\def\predicted#1{#1_{\text{pred}}}

\global\long\def\indexedPredicted#1#2{#1_{\text{pred},\,#2}}


\global\long\def\matrixOfHinton{M}

\global\long\def\mHinton{m}

\global\long\def\row{x}

\global\long\def\dimension{n}


\global\long\def\trainingImage{X}

\global\long\def\reconstructed#1{#1_{\text{rec}}}

\global\long\def\modified#1{#1_{\text{mod}}}

\global\long\def\LTwo{L^{2}}

\global\long\def\length#1{\left\Vert #1\right\Vert }


\global\long\def\FreeShot{\text{F}}

\global\long\def\numberOfFSClasses{N_{c}}

\title{An Effective Hit-or-Miss Layer \\ Favoring Feature Interpretation
as Learned Prototypes Deformations}

\author{A. Deliège, A. Cioppa and M. Van Droogenbroeck\\
University of Liège\\
 Institut Montefiore, Allée de la découverte 10, B-4000 Liège, Belgium\\
 \texttt{\small{}adrien.deliege@uliege.be} }
\maketitle
\begin{abstract}
Neural networks designed for the task of classification have become
a commodity in recent years. Many works target the development of
more effective networks, which results in a complexification of their
architectures with more layers, multiple sub-networks, or even the
combination of multiple classifiers, but this often comes at the expense
of producing uninterpretable black boxes. In this paper, we redesign
a simple capsule network to enable it to synthesize class-representative
samples, called prototypes, by replacing the last layer with a novel
Hit-or-Miss layer. This layer contains activated vectors, called capsules,
that we train to hit or miss a fixed target capsule by tailoring a
specific centripetal loss function. This possibility allows to develop
a data augmentation step combining information from the data space
and the feature space, resulting in a hybrid data augmentation process.
We show that our network, named $\ourModel$, is able to reach better
performances than those reproduced with the initial CapsNet on several
datasets, while allowing to visualize the nature of the features extracted
as deformations of the prototypes, which provides a direct insight
into the feature representation learned by the network\footnote{Supplementary material (codes and videos) may be found \href{https://drive.google.com/drive/folders/17bhe6vLkU-bFNLDDP54FPzGx_xSXsLnW}{here in dedicated folders}.

~}.
\end{abstract}

\section{Introduction}

Convolutional neural networks (CNNs) have become an omnipresent tool
for image classification and have been revolutionizing the field of
computer vision for the last few years. With the emergence of complex
tasks such as ImageNet classification~\cite{Deng2009ImageNet}, the
networks have grown bigger and deeper while regularly featuring new
layers and other extensions. This makes it increasingly difficult
to understand how the networks make their decisions, which led to
the emergence of a new field of research devoted to improve the explainability
of neural networks~(\eg \cite{Zhang2018Interpretable,Zhou2018Interpretable}).
Another usual problem is that CNNs do not generalize well to novel
viewpoints because the spatial relationships between different features
are generally not preserved in CNNs. Therefore, some models were designed
in the spirit of increasing their representational power, hence their
interpretability, by encapsulating information in activated vectors
called capsules, a notion introduced by Hinton in~\cite{Hinton2011Transforming}.

\begin{figure}[t]
\begin{centering}
\includegraphics[width=0.99\columnwidth]{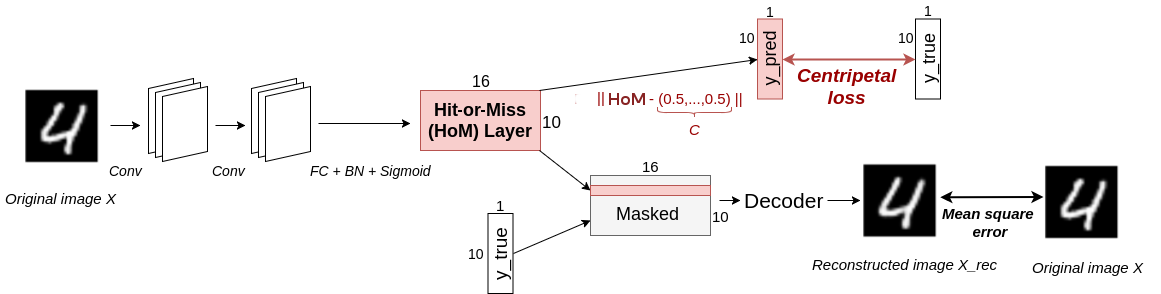}
\par\end{centering}
\caption{Graphical representation of the network structure proposed in this
paper. Our contributions are highlighted in red, and comprise a new
Hit-or-Miss layer, a centripetal loss and prototypes that can be built
with the decoder.}
\label{fig:Hitnet}
\end{figure}

Recent advances on capsules are presented in~\cite{Sabour2017Dynamic},
in which Sabour~\etal mainly focus on MNIST digits classification~\cite{LeCun2001Gradient}.
For that purpose, they develop CapsNet, a CNN that shows major changes
compared to conventional CNNs. As described in~\cite{Sabour2017Dynamic},
``a capsule is a group of neurons whose activity vector represents
the instantiation parameters of a specific type of entity such as
an object or an object part.'' Hence, the concept of capsule somehow
adds a (geometrical) dimension to the ``capsuled'' layers, which
is meant to contain richer information about the features captured
by the network than in conventional feature maps. The transfer of
information from the capsules of a layer to the capsules of the next
layer is learned through a dynamic routing mechanism~\cite{Hinton2018Matrix,Sabour2017Dynamic}.
The length of the capsules of the last layer, called DigitCaps, is
used to produce a prediction vector whose entries are in the $[0,1]$
range thanks to an orientation-preserving squashing activation function
applied beforehand to each capsule, and which encodes the likelihood
of the existence of each digit on the input image. This prediction
vector is evaluated through a ``margin loss'' that displays similarities
with the squared hinge loss. In an autoencoder spirit, the capsules
encoded in DigitCaps are fed to a decoder sub-network that aims at
reconstructing the initial image, which confers the capsules a natural
interpretation of the features that they captured. State-of-the-art
results are reported by Sabour~\etal in~\cite{Sabour2017Dynamic}
on MNIST dataset. Other experiments carried out on affNIST~\cite{Tieleman2013AffNIST},
multiMNIST~\cite{Sabour2017Dynamic}, SVHN~\cite{Netzer2011Reading},
smallNORB~\cite{LeCun2004Learning} and CIFAR10~\cite{Krizhevsky2009Learning}
(with an ensemble of 7 networks) show promising results as well. Unfortunately,
current implementations of CapsNet with dynamic routing are considerably
slower to train than conventional CNNs, which is a major drawback
of this process. 

Since the publication of~\cite{Sabour2017Dynamic}, several works
have been conducted to improve CapsNet's speed and structure \cite{Bahadori2018Spectral,Hinton2018Matrix,Rawlinson2018Sparse,Wang2018AnOptimization}
and to apply it to more complex data \cite{Afshar2018Brain,Li2018TheRecognition,ONeill2018Siamese}
and various tasks: \cite{Liu2018Object} for localization, \cite{Lalonde2018Capsules}
for segmentation,  \cite{Andersen2018DeepReinforcement} for reinforcement
learning. However, it appears that the attempts (\eg \cite{Guo2017CapsNetKeras,Liao2018CapsNet,Nair2018Pushing,Shin2018CapsNetTensorFlow})
to reproduce the results provided in~\cite{Sabour2017Dynamic} failed
to reach the reported performances. 

Our first contribution is the construction of a neural network, named
\emph{HitNet}, that provides fast and repeatedly better performances
than those reported in \cite{Guo2017CapsNetKeras,Liao2018CapsNet,Nair2018Pushing,Shin2018CapsNetTensorFlow}
with reproductions of CapsNet by reusing CapsNet's capsule approach
only in one layer, called \emph{Hit-or-Miss layer} (HoM, the counterpart
of DigitCaps), in a different way. We also provide its associated
loss, that we call \emph{centripetal loss }(counterpart of the margin
loss). Then, we show that plugging the HoM structure into different
existing network architectures consistently improves their classification
results on various datasets. Finally, we present a way of using the
capsules of HoM to derive a hybrid data augmentation algorithm that
relies on both real data and synthetic feature-based data by deforming
\emph{prototypes}, which are class representatives learned indirectly
by the decoder.

\section{Methods}

$\ourModel$ essentially introduces a new layer, the Hit-or-Miss layer,
that is universal enough to be used in many different networks, as
shown in the next section. $\ourModel$ as presented hereafter and
displayed in Figure~\ref{fig:Hitnet} is thus an instance of a shallow
network that hosts this HoM layer and illustrates our point.

\subsection*{The Hit-or-Miss layer\label{subsec:Centripetal-loss-and-HitNet}}

 In the case of CapsNet, large activated values are expected from
the capsule of DigitCaps corresponding to the true class of a given
image, similarly to usual networks. From a geometrical perspective
in the feature space, this results in a capsule that can be seen as
a point that the network is trained to push far from the center of
the unit hypersphere, in which it ends up thanks to the squashing
activation function. We qualify such an approach as ``centrifugal''.
In that case, a first possible issue is that one has no control on
the part(s) of the sphere that will be targeted by CapsNet and a second
one is that the capsules of two images of the same class might be
located far from each other \cite{Shahroudnejad2018Improved,Zhang2018CapProNet},
which are two debatable behaviors. 

To circumvent these potential issues, we impose that all the capsules
of images of a same class should be located close to each other and
in a neighborhood of a given fixed target point. This comes from the
hypothesis that all the images of a given class share some class-specific
features and that this assumption should also hold through their respective
capsules. Hence, given an input image, we impose that $\ourModel$
targets the center of the feature space to which the capsule of the
true class belongs, so that it corresponds to what we call a \emph{hit}.
The capsules related to the other classes have thus to be sent far
from the center of their respective feature spaces, which corresponds
to what we call a\emph{ miss}. Our point of view is thus the opposite
of Sabour \etal's; instead, we have a\emph{ centripetal approach}
with respect to the true class.

Also, the squashing activation function used by Sabour \etal induces
a dependency between the features of a capsule of DigitCaps, in the
sense that their values are conditioned by the overall length of the
capsule. If one feature of a capsule has a large value, then the squashing
prevents the other features of that capsule to take large values as
well; alternatively, if the network wishes to activate many features
in a capsule, then none of them will be able to have a large value.
None of these two cases fit with the perspective of providing strong
activations for several representative features as desired in Sabour
\etal. Besides, the orientation of the capsules, preserved with the
squashing activation, is not used explicitly for the classification;
preserving the orientation might thus be a superfluous constraint.

Therefore, we replace this squashing activation by a BatchNormalization
(BN,~\cite{Ioffe2015BatchNormalization}) followed by a conventional
sigmoid activation function applied element-wise. We obtain a layer
composed of capsules as well that we call the\emph{ Hit-or-Miss} (HoM)
layer, which is $\ourModel$'s counterpart of DigitCaps. Consequently,
all the features obtained in HoM's capsules can span the $[0,1]$
range and they can reach any value in this interval independently
of the other features. The feature spaces in which the capsules of
HoM lie are thus unit hypercubes.

\subsection*{Defining the centripetal loss}

Given the use of the element-wise sigmoid activation, the centers
of the reshaped target spaces are, for each of them, the \emph{central
capsules} $\centerTarget:(0.5,\,\ldots,\,0.5)$. The $\classIndex$-th
component of the prediction vector $\predicted y$ of $\ourModel$,
denoted $\indexedPredicted{\output}{\classIndex}$, is given by the
Euclidean distance between the $\classIndex$-th capsule of HoM and
$\centerTarget$:
\begin{equation}
\indexedPredicted{\output}{\classIndex}=||\text{HoM}_{\classIndex}-\centerTarget||.\label{eq:ourpred}
\end{equation}
To give a tractable form to the notions of hits, misses, centripetal
approach described above and justify HoM's name, we design a custom
centripetal loss function with the following requirements:
\begin{enumerate}
\item The loss generated by each capsule of HoM has to be independent of
the other capsules, which thus excludes any probabilistic notion.
\item The capsule of the true class does not generate any loss when belonging
to a close isotropic neighborhood of $\centerTarget$, which defines
the \emph{hit zone}. Outside that neighborhood, it generates a loss
increasing with its distance to $\centerTarget$. The capsules related
to the remaining classes generate a loss decreasing with their distance
to $\centerTarget$ inside a wide neighborhood of $\centerTarget$
and do not generate any loss outside that neighborhood, which is the
\emph{miss zone}. These loss-free zones are imposed to stop penalizing
capsules that are already sufficiently close (if associated with the
true class) or far (if associated with the other classes) from $\centerTarget$
in their respective feature space.
\item The gradient of the loss with respect to $\indexedPredicted{\output}k$
cannot go to zero when the corresponding capsule approaches the loss-free
zones defined in requirement 2. To guarantee this behavior, we impose
a constant gradient around these zones. This is imposed to help the
network make hits and misses.
\end{enumerate}
For the sake of consistency with requirement 3, in the present work,
we impose piecewise constant gradients with respect to $\indexedPredicted{\output}k$,
which thus defines natural bins around $\centerTarget$, as the rings
of archery targets, in which the gradient is constant. Using smoother
continuous gradient functions did not seem to have any significant
impact.

\begin{figure}
\subfloat[\label{fig:Target-to-hit}]{\centering{}\includegraphics[width=0.47\columnwidth]{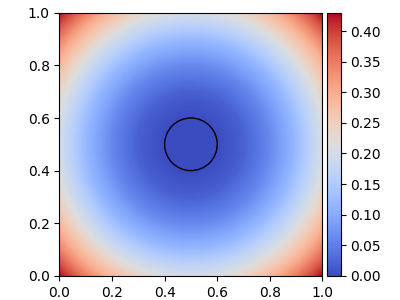}}\hfill{}\subfloat[\label{fig:Target-to-miss}]{\centering{}\includegraphics[width=0.47\columnwidth]{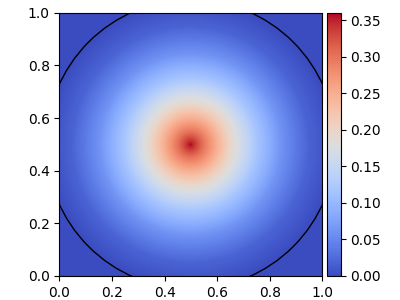}}

\subfloat[]{\centering{}\includegraphics[width=0.47\columnwidth]{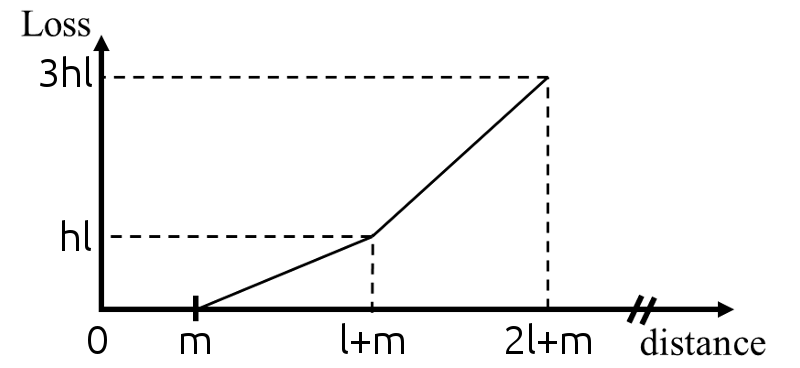}}\hfill{}\subfloat[]{\centering{}\includegraphics[width=0.47\columnwidth]{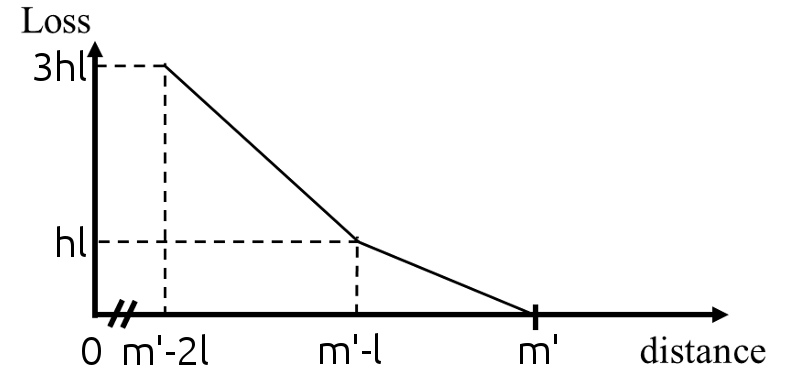}}\caption{Top: Visualization of the centripetal loss in the 2-dimensional case
($\protect\dimension=2$). The loss associated with the capsule of
the true class is given by plot (a). The loss-free hit zone is the
area within the black circle, with radius $m$. The loss generated
by the other capsules is given by plot (b). The loss-free miss zone
is the area outside the black circle, with radius $m'$. Bottom: corresponding
losses expressed as functions of the distance from the centers of
the targets. \label{fig:mylosses} }
\end{figure}

All these elements contribute to define a loss which is a piecewise
linear function of the predictions and which is \emph{centripetal
with respect to the capsule of the true class}. We thus call it our
\emph{centripetal loss}. Its derivative with respect to $\indexedPredicted{\output}k$
is a staircase-like function, which goes up when $k$ is the index
of the true class and goes down otherwise. A generic analytic formula
of a function of a variable $x$, whose derivative is an increasing
staircase-like function where the steps have length $l$, height $h$
and vanish on $[0,m]$ is mathematically given by:

\begin{equation}
L_{l,h,m}(x)=H\{x-m\}\,(f+1)\,h\,(x-m-0.5\,f\,l),\label{eq:steploss1}
\end{equation}

\noindent where $H\{.\}$ denotes the Heaviside step function and
$f=\left\lfloor \frac{x-m}{l}\right\rfloor $ ($\left\lfloor .\right\rfloor $
is the floor function). Such a function is drawn in Figure~\ref{fig:mylosses}.
Hence the loss generated by the capsule of the true class is given
by $L_{l,h,m}(\indexedPredicted{\output}k)$, where $k$ is the index
of the true class. The loss generated by the capsules of the other
classes can be directly obtained from Equation~\ref{eq:steploss1}
as $L_{l',h',\sqrt{n}/2-m'}(\sqrt{n}/2-\indexedPredicted{\output}{k'})$
(for any index $k'$ of the other classes) if the steps have length
$l'$, height $h'$, vanish after $m'$ and if the capsules have $n$
components. The use of $\sqrt{n}/2$ originates from the fact that
the maximal distance between a capsule of HoM and $\centerTarget$
is given by $\sqrt{n}/2$ and thus the entries of $\predicted{\output}$
will always be in the interval $[0,\sqrt{n}/2]$. Consequently, the
centripetal loss of a given training image is given by
\begin{multline}
\loss=\sum_{\classIndex=1}^{\cardinalityOfClasses}\indexedTrue{\output}{\classIndex}\,L_{l,h,m}(\indexedPredicted{\output}{\classIndex})\\
+\lambda(1-\indexedTrue{\output}{\classIndex})\,L_{l',h',\sqrt{n}/2-m'}(\sqrt{n}/2-\indexedPredicted{\output}{\classIndex})\label{eq:myloss-1}
\end{multline}
where $\cardinalityOfClasses$ is the number of classes, $\indexedTrue yk$
denotes the $\classIndex$-th component of the vector $\true y$,
and $\lambda$ is a down-weighting factor set as $0.5$ as in~\cite{Sabour2017Dynamic}.
The loss associated with the capsule of the true class and the loss
associated with the other capsules are represented in Figure~\ref{fig:mylosses}
in the case where $\dimension=2$.

\subsection*{Architecture of $\protect\ourModel$}

Basically, $\ourModel$ incorporates a HoM layer built upon feature
maps and used in pair with the centripetal loss. We have adopted a
shallow structure to obtain these feature maps to highlight the benefits
of the HoM layer. $\ourModel$'s complete architecture is displayed
in Figure~\ref{fig:Hitnet}. First, it has two $9\times9$ (with
strides (1,1) then (2,2)) convolutional layers with $256$ channels
and ReLU activations, to obtain feature maps. Then, it has a fully
connected layer to a $\cardinalityOfClasses\times\dimension$ matrix,
followed by a BN and an element-wise sigmoid activation, which produces
HoM composed of $\cardinalityOfClasses$ capsules of size $\dimension$.
The Euclidean distance with the central capsule $\centerTarget:(0.5,\,\ldots,\,0.5)$
is computed for each capsule of HoM, which gives the prediction vector
of the model $\predicted{\output}$. All the capsules of HoM are masked
(set to $0$) except the one related to the true class (to the predicted
class at test time), then they are concatenated and sent to a decoder,
which produces an output image $\reconstructed{\trainingImage}$,
that aims at reconstructing the initial image. The decoder consists
in two fully connected layers of size $512$ and $1024$ with ReLU
activations, and one fully connected layer to a matrix with the same
dimensions as the input image, with a sigmoid activation (this is
the same decoder as in~\cite{Sabour2017Dynamic}).

If $\trainingImage$ is the initial image and $\true{\output}$ its
one-hot encoded label, then $\true{\output}$ and $\predicted{\output}$
produce a loss $L_{1}$ through the centripetal loss given by Equation~\ref{eq:myloss-1}
while $\trainingImage$ and $\reconstructed{\trainingImage}$ generate
a loss $L_{2}$ through the mean squared error. The final composite
loss associated with $X$ is given by $L=L_{1}+\alpha\,L_{2}$, where
$\alpha$ is set to $0.392$ \cite{Guo2017CapsNetKeras,Sabour2017Dynamic}.
For the classification task, the label predicted by $\ourModel$ is
the index of the lowest entry of $\predicted y$. The hyperparameters
involved in $L_{1}$ are chosen as $l=l'=0.1$, $h=h'=0.2$, $m=0.1$,
$m'=0.9$, $n=16$ and $\lambda=0.5$ as in~\cite{Sabour2017Dynamic}.

\subsection*{Prototypes and hybrid data augmentation\label{subsec:Prototypes,-data-generation-1}}

In our centripetal approach, we ensure that all the images of a given
class will have all the components of their capsule of that class
close to $0.5$. In other words, we regroup these capsules in a convex
space around $\centerTarget$. This central capsule $\centerTarget$
stands for a fixed point of reference, hence different from a centroid,
from which we measure the distance of the capsules of HoM; from the
network's point of view, $\centerTarget$ stands for a capsule of
reference from which we measure deformations.  In consequence, we
can use $\centerTarget$ instead of the capsule of a class of HoM,
zero out the other capsules and feed the result in the decoder: the
reconstructed image will correspond to the image that the network
considers as a canonical image of reference for that class, which
we call its \emph{prototype.}

After constructing the prototypes, we can slightly deform them to
induce variations in the reconstruction without being dependent on
any training image, just by feeding the decoder with a zeroed out
HoM plus one capsule in a neighborhood of $\centerTarget$. This allows
to identify what the features of HoM represent. For the same purpose,
Sabour \etal need to rely on a training image because the centrifugal
approach does not directly allows to build prototypes. In our case,
it is even possible to compute an approximate range in which the components
can be tweaked. If a sufficient amount of training data is available,
we can expect the individual features of the capsules of the true
classes to be approximately Gaussian distributed with mean $0.5$
and standard deviation $m/\sqrt{n}$. This comes from the BN layer
and from Equation~\ref{eq:ourpred}, with the hypothesis that all
the values of such a capsule differ from $0.5$ from roughly the same
amount. Thus the interval $[0.5-2m/\sqrt{n},0.5+2m/\sqrt{n}]$ provides
a satisfying overview of the physical interpretation embodied in a
given feature of HoM.

The capsules of HoM encode deformations of the prototypes, hence they
only capture the important features that allow the network to identify
the class of the images and to perform an approximate reconstruction
via the decoder. This implies that the images produced by the decoder
are not detailed enough to look realistic. The details are lost in
the process; generating them back is hard. It is easier to use already
existing details, \ie those of images of the training set. We can
thus set up a hybrid feature-based and data-based data augmentation
process:
\begin{itemize}
\item Take a training image $\trainingImage$ and feed it to a trained $\ourModel$
network.
\item Extract its HoM and modify the capsule corresponding to the class
of $\trainingImage$.
\item Reconstruct the image obtained from the initial capsule, $\reconstructed{\trainingImage}$,
and from the modified one, $\modified{\trainingImage}$.
\item The details of $\trainingImage$ are contained in $\trainingImage-\reconstructed{\trainingImage}$.
Thus the new (detailed) image is $\modified{\trainingImage}+\trainingImage-\reconstructed{\trainingImage}$.
Clip the values to ensure that the resulting image has values in the
appropriate range (\eg~{[}0,1{]}).
\end{itemize}

\section{Experiments and results\label{sec:Experiments-and-results}}

\subsubsection*{Description of the networks used for comparison.}

First, we compare the performances of $\ourModel$ to three other
networks for the MNIST digits classification task. For the sake of
a fair comparison, a structure similar to $\ourModel$ is used as
much as possible for these networks. First, they are made of two $9\times9$
convolutional layers with $256$ channels (with strides (1,1) then
(2,2)) and ReLU activations as for $\ourModel$. Then, the first network
is N1:
\begin{itemize}
\item N1 (baseline model, conventional CNN) has a fully connected layer
to a vector of dimension $10$, then BN and Softmax activation, and
is evaluated with the usual categorical cross-entropy loss. No decoder
is used.
\end{itemize}
The two other networks, denoted N2 and N2b, have a fully connected
layer to a $10\times16$ matrix, followed by a BN layer as N1 and
$\ourModel$, then
\begin{itemize}
\item N2 (CapsNet-like model) has a squashing activation. The Euclidean
distance with $O:(0,\,\ldots,\,0)$ is computed for each capsule,
which gives the output vector of the model $\predicted{\output}$.
The margin loss (centrifugal) of~\cite{Sabour2017Dynamic} is used;
\item N2b has a sigmoid activation. The Euclidean distance with $\centerTarget:(0.5,\,\ldots,\,0.5)$
is computed for each capsule, which gives the output vector of the
model $\predicted{\output}$. The margin loss (centrifugal) of~\cite{Sabour2017Dynamic}
is used. 
\end{itemize}
Network N2b is tested to show the benefits of the centripetal approach
of $\ourModel$ over the centrifugal one, regardless of the squashing
or sigmoid activations. Also, during the training phase, the decoder
used in $\ourModel$ is also used with N2 and N2b.

\subsubsection*{Classification results on MNIST.}

Each network is trained $20$ times during $250$ epochs with the
Adam optimizer with an initial learning rate of $0.001$, with batches
of $128$ images. The images of a batch are randomly shifted of up
to $2$ pixels in each direction (left, right, top, bottom) with zero
padding as in~\cite{Sabour2017Dynamic}.

First, the learning rate is kept constant to remove its possible influence
on the results. This leads us to evaluate the ``natural'' convergence
of the networks since the convergence is not forced by a decreasing
learning rate mechanism. To our knowledge, this practice is not common
but should be used to properly analyze the natural convergence of
a network.

\begin{figure}
\centering{}\includegraphics[width=0.95\columnwidth]{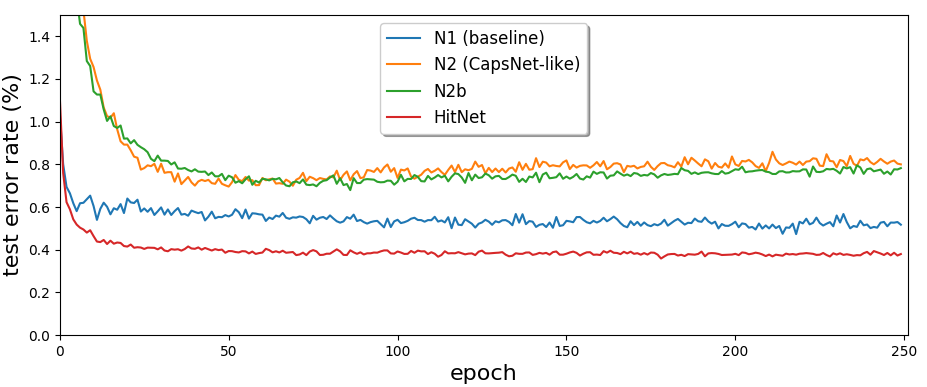}\caption{Evolution of the average test error rate on MNIST over the $20$
runs of each network as a function of the epochs, with a constant
learning rate. The superiority of $\protect\ourModel$ can be seen.
The convergence is also finer in that case, in light of the weaker
oscillations from one epoch to the next.\label{fig:firstcompar-1}}
\end{figure}

The average test error rates per network throughout the epochs are
plotted in Figure~\ref{fig:firstcompar-1}. They clearly indicate
that the centripetal approach of $\ourModel$ is better suited than
a centrifugal approach. It can also be seen that $\ourModel$ does
not suffer from overfitting contrary to N2 and N2b, and that its performance
curve appears more ``regular'' than the others. This indicates that
there is an intrinsically better natural convergence associated with
$\ourModel$'s centripetal approach. 

The average test error rates at the end of the training are reported
in Table~\ref{tab:performances}, confirming the superior performances
of $\ourModel$. The associated standard deviations reveal that more
consistent results between the runs of a same model (lower standard
deviation) are obtained with $\ourModel$. Let us note that all the
runs of $\ourModel$ converged and no overfitting is observed. The
question of the convergence is not studied in \cite{Sabour2017Dynamic}
and the network is stopped before it is observed diverging in \cite{Nair2018Pushing}.
\begin{table}
\centering{}{\small{}}%
\begin{tabular}{|c|c|c|}
\hline 
\multirow{2}{*}{{\small{}Network}} & \multicolumn{2}{c|}{{\small{}Test error rate (\%)}}\tabularnewline
 & {\small{}Const./Decr. learning rate} & {\small{}Best}\tabularnewline
\hline 
{\small{}Baseline} & {\small{}$0.52_{\pm0.060}$$/0.42_{\pm0.027}$} & {\small{}$0.38$}\tabularnewline
\hline 
{\small{}CapsNet-like} & {\small{}$0.79_{\pm0.089}/0.70_{\pm0.076}$} & {\small{}$0.53$}\tabularnewline
\hline 
{\small{}N2b} & {\small{}$0.76_{\pm0.072}/0.72_{\pm0.074}$} & {\small{}$0.62$}\tabularnewline
\hline 
{\small{}$\ourModel$} & {\small{}$\mathbf{0.38_{\pm0.033}/0.36_{\pm0.025}}$} & \textbf{\small{}$\mathbf{0.32}$}\tabularnewline
\hline 
\end{tabular}\caption{Average performance on MNIST test set over the $20$ runs of each
of the four networks with the associated standard deviation, and single
best performance recorded per network.}
\label{tab:performances}
\end{table}

Then, we run the same experiments but with a decreasing learning rate,
to see how the results are impacted when the convergence is forced.
The learning rate is multiplied by a factor $0.95$ at the end of
each epoch. As a result, the networks stabilize more easily around
a local minimum of the loss function, improving their overall performances.
It can be noted from Table~\ref{tab:performances} that $\ourModel$
is less impacted, which indicates that $\ourModel$ converges to similar
states with or without decreasing learning rate. The conclusions are
the same as previously: $\ourModel$ performs better. The best error
rate obtained for a converged run of each type of network is also
indicated in Table~\ref{tab:performances}.

\subsubsection*{Comparisons with reported results of CapsNet on several datasets. }

As far as MNIST is concerned, the best test error rate reported in~\cite{Sabour2017Dynamic}
is $0.25\%$, which is obtained with dynamic routing and is an average
of 3 runs only. However, to our knowledge, the best tentative reproductions
reach error rates which compare with our results, as shown in Table~\ref{tab:compar}.
It is important to underline that such implementations report excessively
long training times, mainly due to the dynamic routing part. For example,
the implementation~\cite{Guo2017CapsNetKeras} appears to be about
$13$ times slower than $\ourModel$, for comparable performances.
Therefore, $\ourModel$ produces results consistent with state-of-the
art methods on MNIST while being simple, light and fast. For the record,
in~\cite{Wan2013Regularization}, authors report a $0.21\%$ test
error rate, which is the best performance published so far. Nevertheless,
this score is reached with a voting committee of five networks that
were trained with random crops, rotation and scaling as data augmentation
processes. They achieve $0.63\%$ without the committee and these
techniques; if only random crops are allowed (as done here with the
shifts of up to 2 pixels), they achieve $0.39\%$.

The results obtained with $\ourModel$ and those obtained with CapsNet
in different sources on Fashion-MNIST~\cite{Xiao2017Fashion}, CIFAR10,
SVHN, affNIST are also compared in Table~\ref{tab:compar}. The architecture
of $\ourModel$ described previously is left untouched and the corresponding
results reported are obtained with a constant learning rate (excepted
for MNIST) and are average test error rates on $20$ runs as previously.
Some comments about these experiments are given below:
\begin{enumerate}
\item Fashion-MNIST: $\ourModel$ outperforms reproductions of CapsNet except
for~\cite{Guo2017CapsNetKeras}, but this result is obtained with
horizontal flipping as additional data augmentation process. 
\item CIFAR10: $\ourModel$ outperforms the reproductions of CapsNet. The
result provided in~\cite{Sabour2017Dynamic} is obtained with an
ensemble of $7$ models. However, the individual performances of $\ourModel$
and of the reproductions do not suggest that ensembling them would
lead to that result, as also suggested in~\cite{Xi2017Capsule},
reporting between $28\%$ and $32\%$ test error rates.
\item SVHN: $\ourModel$ outperforms CapsNet from~\cite{Nair2018Pushing},
which is the only source using CapsNet with this dataset.
\item affNIST: $\ourModel$ outperforms the results provided in~\cite{Shin2018CapsNetTensorFlow}
and even in Sabour \etal by a comfortable margin. Each image of the
MNIST train set is placed randomly (once and for all) on a black background
of $40\times40$ pixels, which constitutes the training set of the
experiment, and data augmentation is forbidden.. After training, the
models are tested on affNIST test set, which consists in affine transformations
of MNIST test set. 
\end{enumerate}
\begin{table}
\centering{}{\small{}}%
\begin{tabular}{|c|c|c|c|c|c|}
\cline{2-6} 
\multicolumn{1}{c|}{} & {\footnotesize{}MNIST} & {\footnotesize{}Fash.} & {\footnotesize{}CIF10} & {\footnotesize{}SVHN} & {\footnotesize{}aff.}\tabularnewline
\hline 
{\footnotesize{}Sabour}{\small{} }\emph{\small{}et al.} & {\small{}0.25} & {\small{}-} & {\small{}10.60} & {\small{}4.30} & {\small{}21.00}\tabularnewline
\hline 
{\footnotesize{}Nair}\emph{\small{} et al.} & {\small{}0.50} & {\small{}10.20} & {\small{}32.47} & {\small{}8.94} & {\small{}-}\tabularnewline
\hline 
{\footnotesize{}Guo} & {\small{}0.34} & {\small{}6.38} & {\small{}27.21} & {\small{}-} & {\small{}-}\tabularnewline
\hline 
{\footnotesize{}Liao} & {\small{}0.36} & {\small{}9.40} & {\small{}-} & {\small{}-} & {\small{}-}\tabularnewline
\hline 
{\footnotesize{}Shin} & {\small{}0.75} & {\small{}10.98} & {\small{}30.18} & {\small{}-} & {\small{}24.11}\tabularnewline
\hline 
\hline 
{\footnotesize{}$\ourModel$} & {\small{}0.36} & {\small{}7.70} & {\small{}26.70} & {\small{}5.50} & {\small{}16.97}\tabularnewline
\hline 
\end{tabular}\caption{Comparison between the test error rates (in \%) reported on various
experiments with CapsNet and $\protect\ourModel$, in which case the
average results over $20$ runs are reported.}
\label{tab:compar}
\end{table}

\subsubsection*{HoM with other encoder architectures. }

The use of the HoM layer is not restricted to $\ourModel$, as it
can be easily plugged into any network architecture. In this section,
we compare the performances obtained on MNIST, Fashion-MNIST, CIFAR10
and SVHN with and without the HoM when the encoder sub-network corresponds
to well-known architectures: LeNet-5 \cite{LeCun2001Gradient}, ResNet-18
\cite{He2015DeepResidual} and DenseNet-40-12 \cite{Huang2017Densely}.
Incorporating the HoM consists in replacing the last fully connected
layer (with softmax activation) to a vector of size $\cardinalityOfClasses$
by a fully connected layer to a $\cardinalityOfClasses\times\dimension$
matrix, followed by a BN and an element-wise sigmoid activation, which
produces HoM composed of $\cardinalityOfClasses$ capsules of size
$\dimension$. We keep the exact same structure as the one described
for $\ourModel$. As previously, for each dataset, each network is
trained from scratch $20$ times during $250$ epochs, with a constant
learning rate and with random shifts of up to $2$ pixels in each
direction as sole data augmentation process.

The relative decrease in the average test error rates when passing
from models without HoM to models with HoM are reported in Table~\ref{tab:perfs-other-netw-1}.
The decrease computed from the previous metrics for N1 versus $\ourModel$
is also reported. As it can be seen in Table~\ref{tab:perfs-other-netw-1},
using the HoM layer in a network almost always decreases the test
error rates, sometimes by a comfortable margin ($>15\%$). As in Figure~\ref{fig:firstcompar-1},
we noticed that the learning curves using the HoM appear much smoother
than those from the initial networks. We also noticed that the average
standard deviation in the $20$ error rates per model is generally
lower ($13$ experiments among $16$) when using the HoM. This metric
decreases by at least $20\%$ (and up to $70\%$) for $9$ experiments
out of $16$, which implies that the use of HoM and the centripetal
loss may help producing more consistent results between several runs
of a same network.
\begin{table}
\begin{centering}
{\small{}}%
\begin{tabular}{|c|c|c|c|c|}
\cline{2-5} 
\multicolumn{1}{c|}{} & {\small{}MNIST} & {\small{}Fash.} & {\small{}CIF10} & {\small{}SVHN}\tabularnewline
\hline 
{\small{}N1} & {\small{}26.9} & {\small{}8.9} & {\small{}0.9} & {\small{}25.5}\tabularnewline
\hline 
{\small{}LeNet-5} & {\small{}17.3} & {\small{}5.3} & {\small{}5.9} & {\small{}3.9}\tabularnewline
\hline 
{\small{}ResNet-18} & {\small{}20.8} & {\small{}7.5} & {\small{}8.1} & {\small{}17.2}\tabularnewline
\hline 
{\small{}DenseNet-40-12} & {\small{}6.4} & {\small{}5.1} & {\small{}-0.7} & {\small{}15.1}\tabularnewline
\hline 
\end{tabular}{\small\par}
\par\end{centering}
\caption{Relative decrease (in \%) in the average test error rates when using
the HoM with several architectures.}
\label{tab:perfs-other-netw-1}
\end{table}

\subsubsection*{Examining prototypes. }

The centripetal approach gives a particular role to the central capsule
$\centerTarget:(0.5,...,0.5)$, in the sense that it can be fed to
the decoder to generate prototypes of the different classes. The prototypes
obtained from an instance of $\ourModel$ trained on MNIST are displayed
in Figure~\ref{fig:prototypes}. Videos of their evolution through the epochs can be found in supplementary material.

\begin{figure}
\begin{centering}
\includegraphics[width=0.95\columnwidth]{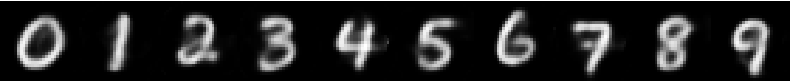}
\par\end{centering}
\centering{}\caption{Prototypes obtained at the end of the training by feeding the decoder
with capsules of zeros except one, which is replaced by the central
capsule $\protect\centerTarget:(0.5,\,\ldots,\,0.5)$. These prototypes
can be seen as the reference images from which $\protect\ourModel$
evaluates the similarity with the input image through HoM. \label{fig:prototypes}}
\end{figure}

\begin{figure}
\begin{centering}
\includegraphics[width=0.95\columnwidth]{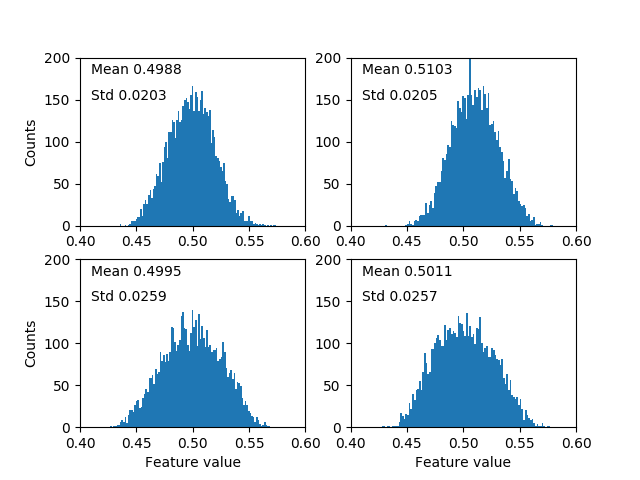}
\par\end{centering}
\centering{}\caption{Distributions of the values of some individual features of the capsule
``4'' among all the images of the digit ``4''. The centripetal
approach allows to obtain Gaussian-like distributions with mean $\approx0.5$
and standard deviation $\approx0.025$ as expected.}
\label{fig:histogs}
\end{figure}
\begin{figure}
\begin{centering}
\includegraphics[width=0.95\columnwidth]{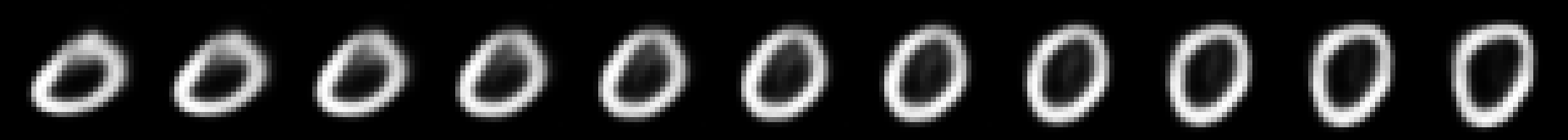}
\par\end{centering}
\begin{centering}
\includegraphics[width=0.95\columnwidth]{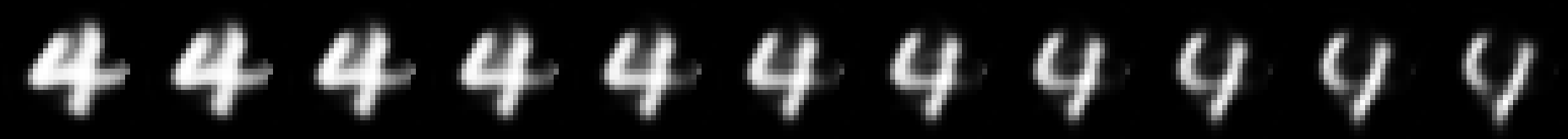}
\par\end{centering}
\begin{centering}
\includegraphics[width=0.95\columnwidth]{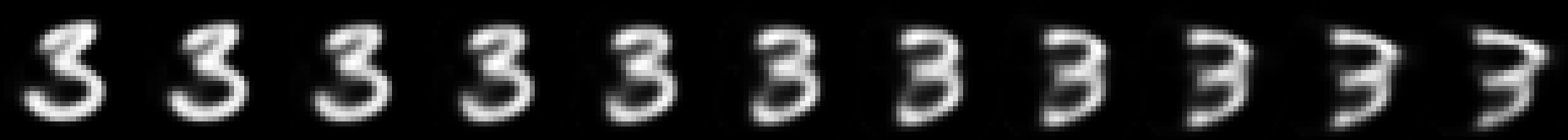}
\par\end{centering}
\centering{}\caption{Visualization of some features captured in the capsules. The prototypes
(in the middle column) are deformed by variations in one entry between
$0.45$ and $0.55$ (row-wise). Different kinds of features appear
in the deformations of the prototypes.}
\label{fig:features}
\end{figure}

As can be seen in Figure~\ref{fig:histogs}, the distributions of
the individual features of the capsules of the true classes (\eg
capsule 4 for images of the digit ``4'') are approximately Gaussian
distributed with mean $0.5$ and standard deviation $0.1/\sqrt{16}=0.025$
as explained previously. This confirms that the interval $[0.5-2*0.025,0.5+2*0.025]$
directly provides a satisfying range to visualize the physical interpretation
embodied in a given feature of HoM. This is illustrated in Figure~\ref{fig:features},
where one feature of the prototype (in the middle column of each row)
is tweaked between $0.45$ and $0.55$ by steps of $0.01$. In our
case, in order to visualize the nature of a feature, there is no need
to distort real images as in~\cite{Sabour2017Dynamic}. As seen
in Figure~\ref{fig:features}, on MNIST, $\ourModel$ captures some
features that are positional or scale characteristics, others can
be related to the width of the font or to some local peculiarities
of the digits, as in~\cite{Sabour2017Dynamic}. Sampling random capsules
close to $\centerTarget$ for a given class thus generates new images
whose characteristics are combinations of the characteristics of the
training images. It thus makes sense to encompass all the capsules
of training images of that class in a convex space, as done with $\ourModel$,
to ensure the consistency of the images produced, while CapsNet does
not guarantee this behavior. Let us note that, as underlined in~\cite{Nair2018Pushing},
the reconstructions obtained for Fashion-MNIST lacks details and those
of CIFAR10 and SVHN are somehow blurred backgrounds; this is also
the case for the prototypes. We believe that at least three factors
could provide an explanation: the decoder is too shallow, the size
of the capsules is too short, and the fact that the decoder has to
reconstruct the whole image, including the background, which is counterproductive.

\subsubsection*{Hybrid data augmentation. }

\begin{figure}
\subfloat[]{\includegraphics[width=0.47\columnwidth]{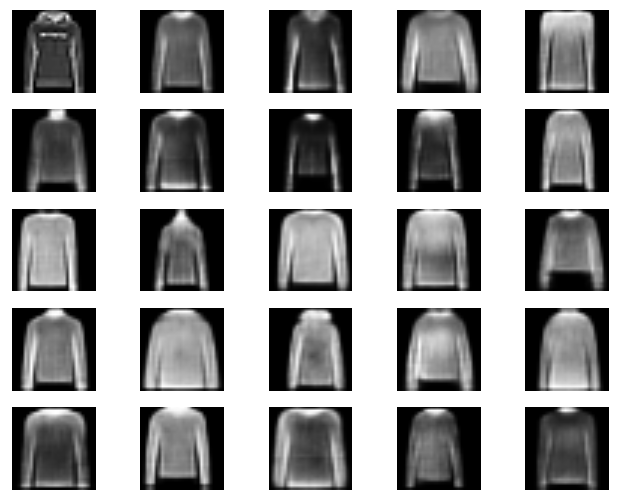}}\hfill{}\subfloat[]{\includegraphics[width=0.47\columnwidth]{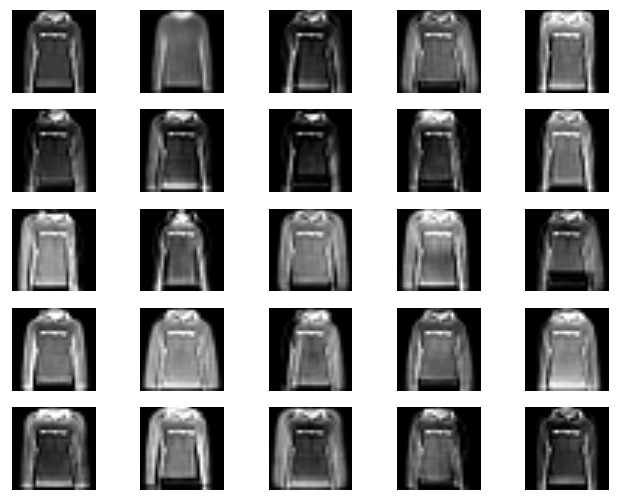}}

\caption{Examples of hybrid data augmentation using the Fashion-MNIST dataset.
In (a) and (b), the top left image is an original image $\protect\trainingImage$
from the dataset and the image on its right is its corresponding reconstruction
$\protect\reconstructed{\protect\trainingImage}$ from HoM (these
images are the same in (a) and (b)). In (a), the $23$ remaining images
are modified versions of $\protect\reconstructed{\protect\trainingImage}$
obtained by tweaking the components of its capsule in HoM. These images
correspond to $23$ examples of $\protect\modified{\protect\trainingImage}$.
Adding the details $\protect\trainingImage-\protect\reconstructed{\protect\trainingImage}$
to each of them gives the $23$ remaining images of (b), which are
the images generated by the hybrid data augmentation explained in
the text.\label{fig:dataaug}}
\end{figure}

In order to incorporate the details lost in the computation of HoM,
the hybrid feature-based and data-based data augmentation technique
can be applied. During the step consisting in a modification of the
capsules, their values are tweaked  of at most $0.025$. The importance
of adding the details and thus the benefits over the sole data generation
process can be visualized in Figure~\ref{fig:dataaug} with the Fashion-MNIST
dataset. 

The classification performances can be marginally increased with this
data augmentation process as it appeared that networks trained from
scratch on such data (continuously generated on-the-fly) perform slightly
better than when trained with the original data. On MNIST, the average
error rate on $20$ models decreased to $0.33\%$ with a constant
learning rate and to $0.30\%$ with a decreasing learning rate. In
our experiments, $3$ of these models converged to $0.26\%$, one
converged to $0.24\%$. Some runs reached $0.20\%$ test error rate
at some epochs. With a bit of luck, a blind selection of one trained
network could thus lead to a new state of the art, even though it
is known that MNIST digits classification results cannot reach better
performances due to inconsistencies in the test set, hence the importance
of averaging over many runs to report more honest results. The average
test error rate on Fashion-MNIST decreases by $0.2\%$, on CIFAR10
by $1.5\%$ and on SVHN by $0.2\%$. These results could presumably
be improved with more elaborate encoders and decoders, given the increased
complexity of these datasets.

\section{Conclusion}

We introduce $\ourModel$, a deep learning network characterized by
the use of a Hit-or-Miss layer composed of capsules, which are compared
to central capsules through a new centripetal loss. The idea is that
the capsule corresponding to the true class has to make a hit in its
target space, and the other capsules have to make misses. The novelties
reside in the reinterpretation and in the use of the HoM layer, which
provides new insights on feature interpretability through deformations
of prototypes, which are class representatives. These can be used
to perform data generation and to set up a hybrid data augmentation
process, which is done by combining information from the data space
and from the feature space.

In our experiments, we demonstrate that $\ourModel$ is capable of
reaching state-of-the-art performances on MNIST digits classification
task with a shallow architecture and that it outperforms the results
reproduced with CapsNet on several datasets, while being at least
$10$ times faster to train and facilitating feature interpretability.
The convergence of $\ourModel$ does not need to be forced by a decreasing
learning rate mechanism to reach good performances. $\ourModel$ does
not seem to suffer from overfitting, and provides a small variability
in the results obtained from several runs. We also show that inserting
the HoM into various network architectures allows to decrease their
error rates, which underlies the possible universal use of this layer.
It also appears that the results are generally more consistent between
several runs of a network when using the HoM, and that the learning
curves are more regular. Finally, we show how prototypes can be built
as class representatives and we illustrate the hybrid data augmentation
process to generate new realistic data. This process can also be used
to marginally increase classification performances.

\subsubsection*{Future work. }

The prototypes and all the reconstructions made by the decoder could
be improved by using a more advanced decoder sub-network and capsules
with more components. In real-life cases such as CIFAR10 and SVHN,
it could also be useful to make a distinction between the object of
interest and the background. For example, features designed to reconstruct
the background only could be used. If segmentation masks are available,
one could also use the capsules to reconstruct the object of interest
in the segmented image, or simply the segmentation mask. One could
also imagine to attach different weights to the features captured
by the capsules, so that those not useful for the classification are
used in the reconstruction only. The flexibility of HoM allows to
implement such ideas easily.

\subsubsection*{Acknowledgements}

We are grateful to M. Braham for introducing us to the work of Sabour
et al. and for all the fruitful discussions. This research is supported
by the DeepSport project of the Walloon region, Belgium. A. Cioppa
has a grant funded by the FRIA, Belgium.

\bibliographystyle{aaai}

\end{document}